\documentclass[9pt,conference]{IEEEtran}
\IEEEoverridecommandlockouts
\usepackage{cite}
\usepackage{amsmath,amssymb,amsfonts}
\usepackage{algorithmic}
\usepackage{graphicx}
\usepackage{textcomp}
\usepackage{xcolor}
\def\BibTeX{{\rm B\kern-.05em{\sc i\kern-.025em b}\kern-.08em
    T\kern-.1667em\lower.7ex\hbox{E}\kern-.125emX}}

\usepackage{physics}
\usepackage{booktabs}
\usepackage{multirow}
\usepackage{makecell} 
\usepackage{adjustbox}
\usepackage{multirow}
\usepackage{algorithm}
\usepackage[capitalise]{cleveref}
\usepackage{subcaption}
\newcommand{\changes}[1]{{\color{black}#1}}

\title{ZOQO: Zero-Order Quantized Optimization
\thanks{This research was supported by the Israeli Innovation Authority through the Trust.AI consortium, the TAD center at Tel Aviv University, and KLA.}}
\author{\IEEEauthorblockN{Noga Bar}
\IEEEauthorblockA{\textit{Electrical Engineering} \\
\textit{Tel Aviv University}\\
}
\and
\IEEEauthorblockN{Raja Giryes}
\IEEEauthorblockA{\textit{Electrical Engineering} \\
\textit{Tel Aviv University}\\
}}

\begin{document}

\maketitle
\begin{abstract}
The increasing computational and memory demands in deep learning present significant challenges, especially in resource-constrained environments.
We introduce a zero-order quantized optimization (ZOQO) method designed for training models with quantized parameters and operations.
Our approach leverages zero-order approximations of the gradient sign and adapts the learning process to maintain the parameters' quantization without the need for full-precision gradient calculations.
We demonstrate the effectiveness of ZOQO through experiments in fine-tuning of large language models and black-box adversarial attacks. Despite the limitations of zero-order and quantized operations training, our method achieves competitive performance compared to full-precision methods, highlighting its potential for low-resource environments. 
\end{abstract}
\section{Introduction}
\label{sec:intro}
The optimization and deployment of deep learning models demand significant computational and memory resources.
During the optimization phase, first-order gradient information is typically computed via backward passes, contributing to the high computational costs.
Additionally, the storage requirements for these large scale models are considerable, especially when working with full-precision parameters.
These challenges are pressing as models are deployed in resource-constrained environments, such as edge devices or mobile applications, where both computational power and memory are limited. 

Prior works addressed the first problem by proposing zero-order (ZO) optimization techniques that train deep networks without using the computationally exhaustive back-propagation and are applicable for the scenarios where gradients are not available. Other efforts faced the second problem by using quantization. However, none have addressed both challenges together. 

In this work, we propose a novel method that enables training models in settings with limited computational and memory resources by combining ZO optimization with quantized training. Our proposed Zero-Order Quatized Optimization (ZOQO) framework eliminates the need for full precision calculations for first-order gradients and parameter updates.
This reduces the computational burden and allows the optimization process to be performed using quantized operations, thus minimizing costly memory usage.

ZOQO utilizes the Zero-Sign Stochastic Gradient Descent (ZO-SignSGD) \cite{liu2019signsgd} method.
ZO-SignSGD estimates the sign of the gradients through forward passes (without back-propagation) and updates the parameters with a uniform magnitude regardless of the magnitude of their true gradient.
We adapt this approach to the quantization setting in two ways.
The first is at the sign estimation step. We inject quantized noise into the parameters instead of the normally distributed noise, which is not discrete.
The second is scaling the learning rate according to the quantization scale which enables maintaining the parameters quantized.
As a result, all updates throughout the optimization process are carried out in a quantized format, making our method highly suitable for resource-constrained settings.

To validate the effectiveness of our approach, we employed ZOQO within black-box adversarial attacks that use sign-based ZO optimization. We demonstrate that the failure rates of these attacks, when trained in a quantized environment, suffer minor degradations compared with Full Precision (FP) attacks on quantized models and attacks. 
This exposes the vulnerability of deep models to black-box adversarial attacks in a low-resource setting.

Furthermore, we fine-tuned Large-Language-Models (LLMs) for sentiment analysis with LoRA under low-resource conditions using our framework. We observe a minor degradation for high bit budgets and non-trivial performance for aggressive quantization.
The results highlight the potential of ZOQO for enabling model training mechanisms in constrained low-resource settings.

\begin{figure*}[htbp]
\vspace{-0.5cm}
    \centering
    \begin{subfigure}[b]{0.35\textwidth}
        \centering
        \includegraphics[width=\linewidth]{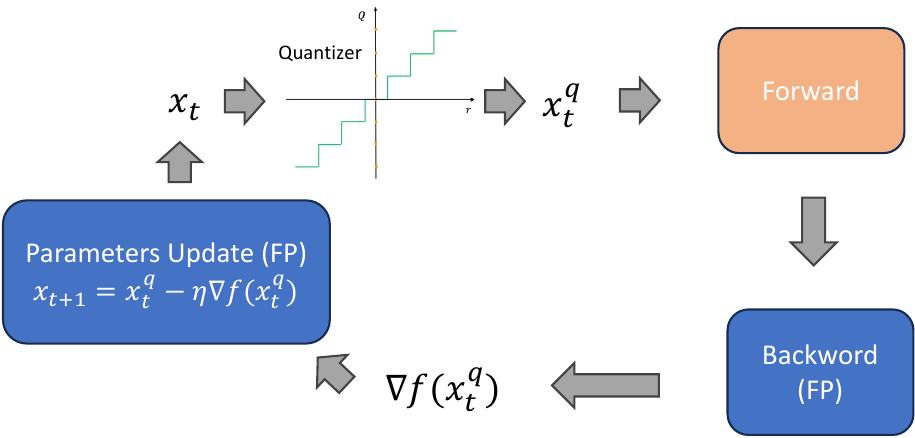}
        \vspace{-0.1cm}
        \caption{QAT - Quantization Aware Training.}
        \label{fig:subfig1}
    \end{subfigure}
    \hspace{0.5cm}
    \begin{subfigure}[b]{0.35\textwidth}
        \centering
        \includegraphics[width=\linewidth]{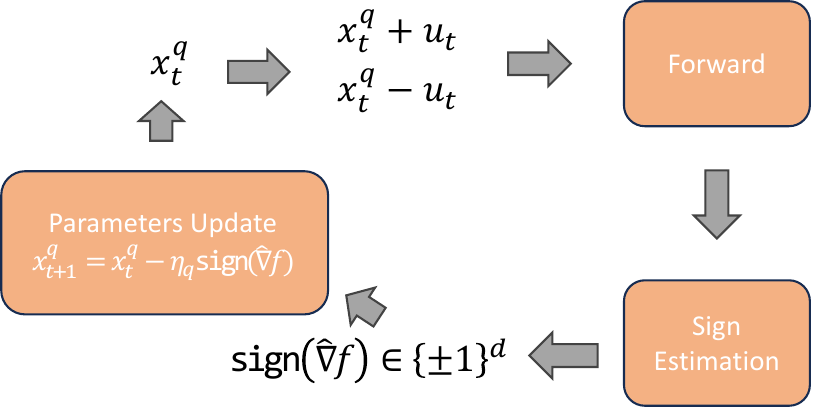}
        \vspace{-0.1cm}
        \caption{ZOQO - Zero Order Quantized Optimization.}
        \label{fig:subfig2}
    \end{subfigure}
    \vspace{-0.2cm}
    \caption{\changes{QAT and ZOQO update steps. 
    Unlike QAT, ZOQO does not use FP operations, making it efficient for resource-constrained setting.}}
    \label{fig:main_figure}
    \vspace{-0.5cm}
\end{figure*}

\section{Related Work}
\textbf{ZO Optimization} methods are backpropagation-free and gained significant attention due to their applicability in scenarios where gradient information is unavailable or impractical to compute \cite{liu2020primer}.
The foundational works of \cite{flaxman2005online,ghadimi2013stochastic,duchi2015optimal,nesterov2017random} laid the groundwork for these methods, establishing their convergence.
Subsequent research is focused on enhancing the efficiency and applicability of zero-order methods in various settings, such as large-scale models \cite{chendeepzero,shu2023zeroth,guo2024zerothorder} including fine tuning of LLMs \cite{malladi2023fine,zhang2024revisiting,gautamvariance}, variance reduction technique \cite{liu2018zeroth}, and memory-efficient training in the general case \cite{cai2021zeroth} and for optical neural networks \cite{gu2021efficient}.
Others, reduced the dependence on the dimension of the convergence rate \cite{wang2018stochastic,balasubramanian2018zeroth,cai2022zeroth,cheng2021convergence}.
Zero-order optimization is widely explored in the context of adversarial attacks \cite{chen2017zoo,ilyas2018black,tu2019autozoom,zhao2019design} and defenses \cite{zhangrobustify} in the black box setting where there is no access to the gradients.
None of the above considered ZO in the quantized setting. 

\noindent \textbf{Quantization} reduces the computational and memory requirements of models by representing weights and activations with lower precision. In this work, we focus on parameter quantization.
Previous quantization-aware training methods, such as those in \cite{jacob2018quantization} and reviewed in \cite{gholami2022survey}, ensure models are quantized during training but typically rely on full-precision operations for gradient calculations.
Methods like DoReFa \cite{zhou2016dorefa} perform quantized training but still require higher precision for gradient calculations.
Extreme quantization techniques for binary or ternary networks, use the sign of the weights for quantization but update the weights with floating-point operations before quantizing them again \cite{darabi2018bnn+, gong2019differentiable, liu2018bi, courbariaux2015binaryconnect, hubara2016binarized, rastegari2016xnor}.
Our work introduces a fully quantized zero-order optimization framework, using the sign of the gradients to maintain parameter quantization.

\begin{algorithm}[t] \caption{ZOQO: Zero-Order Quantized Optimization}\label{algorithm} 
\begin{algorithmic}[1] 
\STATE \textbf{Input:} LR $\eta$, initial parameters $\vb{x_0}$, bit budget $b$, ZO step size $\mu$ 
\STATE \textbf{Initial Quantization:} Quantization scale: $s = \frac{\vb{x_0}_{\max} - \vb{x_0}_{\min}}{2^b-1}$
\STATE Quantize initial parameters: $\vb{x}_0^q = Q(\vb{x}_0)$ (\cref{eq:quantization}) \STATE Compute quantized learning rate: $\eta_q = c(\eta)$ (\cref{eq:lr})
\FOR {$i=1,...,T$} 
\STATE Sample quantized noise: $\vb{u}_i \in \mathcal{B}^d$ 
\STATE Noise injection: 
\vspace{-0.2cm}
\begin{align*}\vb{x_i}^+ = \underset{R_{\min}, R_{\max}}{\mathrm{clamp}} (\vb{x_i}+\vb{u_i}), \hspace{0.2cm} \vb{x_i}^- = \underset{R_{\min}, R_{\max}}{\mathrm{clamp}} (\vb{x_i}-\vb{u_i})
\end{align*}
\STATE Estimate the gradient's sign:
\begin{align*} \mathrm{sign}\left(\hat{\nabla}f \right) = \mathrm{sign}\Big(\ell(\vb{x_i}^+) - \ell(\vb{x_i}^-)\Big) \mathrm{sign}\Big(\vb{u_i}\Big) \end{align*}
\STATE Update parameters:
\vspace{-0.2cm}
\begin{align*}
    \vb{x_i} = \underset{R_{\min}, R_{\max}}{\mathrm{clamp}}\left(\vb{x_{i-1}} - \eta_q \cdot \mathrm{sign}\left(\hat{\nabla}f \right)\}\right)
\end{align*}
\ENDFOR \end{algorithmic}
\end{algorithm}

\section{Quantized Learning}\label{sec:method}
Our quantized learning method is based on ZO-optimization, wherein multiple queries are employed to estimate the gradient of the parameters.
A widely-used technique for ZO is the Randomized Gradient Estimation (RGE) approach \cite{nesterov2017random,ghadimi2013stochastic,duchi2015optimal}.
In this approach, noise is added to the learnable parameters, and the resulting perturbed losses are averaged to estimate the gradient: 
\vspace{0.2cm}
\begin{align} \label{RGE} \hat{\nabla}f = \frac{1}{q} \sum_{i=1}^q \frac{\ell(\vb{x}+\mu \vb{u_i}) - \ell(\vb{x} - \mu \vb{u_i})}{2\mu} \vb{u_i} , \end{align} 
\vspace{0.1cm}
where $\vb{u_i}$ are random vectors usually sampled from a Gaussian distribution.
Note that in \cref{RGE}, the addition of the noise to the parameters, the queries of the model, the calculation of $\ell(\cdot)$ and the average require full precision operations.

\begin{table*}
\centering
\vspace{-0.3cm}
\caption{Failure rate of black-box adversarial attacks 
with different quantization levels. RandSign and SignHunter are zero-order sign-based attack methods that we apply with different quantization strategies. 
On the left are naturally trained models, and on the right are robustly trained models. Multiple quantization settings are compared: Full Precision (FP), quantized models attacked with FP, post-quantized models where attacks are learned in FP and then quantized, and our method, ZOQO.
 Lower is better.
}\label{tab:adv_cifar_mnist}
\begin{tabular}{lllcccc|cccc}
\toprule
\textbf{Dataset} & \textbf{Bits} & \textbf{Method} & \textbf{FP} & \makecell{\textbf{Quant.} \\ \textbf{Model}} & \makecell{\textbf{Post} \\ \textbf{Quant.}} & \makecell{\textbf{ZOQO}} & \textbf{FP} & \makecell{\textbf{Quant.} \\ \textbf{Model}} & \makecell{\textbf{Post}\\ \textbf{Quant.}} & \makecell{\textbf{ZOQO}}  $\downarrow$ \\ \midrule

\multirow{4}{*}{MNIST} & \multirow{2}{*}{$b=8$} 
    & RandSign        & 0.32 & 0.32 & 0.33 & 0.35 & 0.94 & 0.93 & 0.93 & 0.93 \\
    & & SignHunter      & 0.00 & 0.00 & 0.00 & 0.00 & 0.93 & 0.92 & 0.89 & 0.89 \\ \cmidrule{3-11}

& \multirow{2}{*}{$b=4$} 
    & RandSign        & 0.32 & 0.31 & 0.30 & 0.74 & 0.94 & 0.07 & 0.08 & 0.00 \\
    & & SignHunter      & 0.00 & 0.00 & 0.00 & 0.00 & 0.93 & 0.09 & 0.07 & 0.00 \\ \midrule

\multirow{4}{*}{CIFAR-10} & \multirow{2}{*}{$b=8$} 
    & RandSign        & 0.20 & 0.21 & 0.21 & 0.25 & 0.95 & 0.95 & 0.94 & 0.95 \\
    & & SignHunter      & 0.07 & 0.08 & 0.08 & 0.09 & 0.53 & 0.53 & 0.53 & 0.56 \\ \cmidrule{3-11}

& \multirow{2}{*}{$b=4$} 
     & RandSign        & 0.20 & 0.20 & 0.16 & 0.22 & 0.95 & 0.90 & 0.88 & 0.90 \\
    & & SignHunter      & 0.07 & 0.06 & 0.03 & 0.02 & 0.53 & 0.32 & 0.33 & 0.57 \\
\bottomrule
\end{tabular}
\vspace{-0.5cm}
\end{table*}

\begin{table}[t]
    \centering
\caption{Failure rate of black-box adversarial attacks on ImageNet with 8-bit / 4-bit quantization, respectively.}
    \label{tab:adv_imagenet}
    \begin{tabular}{lccc}
    \toprule
         \textbf{Method} & \textbf{FP} & \textbf{Post quant.} & \textbf{ZOQO} $\downarrow$\\ 
         \midrule
        RandSign & 0.72 & 0.73 / 0.67 & 0.73 / 0.67  \\ 
        SignHunter & 0.02 & 0.02 / 0.0 & 0.03 / 0.0 \\ \bottomrule
    \end{tabular}
    \vspace{-0.6cm}
\end{table}

To handle these challenges, instead of using \cref{RGE} for gradient estimations and updates, we built on the ZO-Sign SGD \cite{liu2019signsgd}, where only the sign of the estimated gradients is used to update the model parameters.
The same magnitude of update is applied to every coordinate of the trainable parameters.
The update step of the parameters between two consecutive iterations is:
\changes{
$\vb{x_{t+1}} - \vb{x_{t}} = -\eta \mathrm{sign}(\hat{\nabla}f), $
}
where $\eta$ is the learning rate.
We selected this method because its fixed step size is well suited for quantized parameter updates.
Note that in the original ZO-Sign SGD method, the update steps are not designed to use quantized steps.

We quantize the learned parameters at initialization using uniform quantization with a budget of $b$ bits.
Let $[R_{\min}, R_{\max}]$ be the range of values to be quantized. The quantization operator is defined as: 
\begin{align}\label{eq:quantization}
s = \frac{R_{\max} - R_{\min}}{2^b - 1} , \quad Q(r) = \mathrm{int}\left(\frac{r}{s}\right). \end{align}
Note that although we use uniform quantization for simplicity, more sophisticated methods could be applied.

Our method, ZOQO, modifies the ZO-Sign SGD method at two key points to enable quantized training. The complete algorithm is detailed in \cref{algorithm}.
The first adaptation involves replacing the injected Gaussian noise with quantized noise.
Let $m = \max\{\lfloor\frac{\mu}{s}\rfloor, 1\}$, where $m$ represents the maximum number of discrete quantized steps that can be added to or subtracted from the parameters.
At each training step $i$, the noise is sampled element-wise from the set $\mathcal{B} = \{-ms, (-m+1)s,\dots, 0,\dots, (m-1)s, ms\}$, i.e., $\vb{u_i} \in \mathcal{B}^d$.
This implies that the noise injection step is simply an addition operation of two quantized vectors.
We estimate the sign of the gradient using parameters perturbed by the quantized noise according to the RGE formula in \cref{RGE} with a single noise vector, i.e., $q=1$.
\begin{align*} 
\vb{x}^+ = \underset{R_{\min}, R_{\max}}{\mathrm{clamp}} (\vb{x}+\vb{u_i}), \quad \vb{x}^- = \underset{R_{\min}, R_{\max}}{\mathrm{clamp}} (\vb{x}-\vb{u}), \end{align*} 
\begin{align*}
\mathrm{sign}\left(\hat{\nabla}f \right) = \mathrm{sign} \Big(\ell(\vb{x}^+) - \ell(\vb{x}^-)\Big) \mathrm{sign}\Big(\vb{u}\Big).
\end{align*} 
Note that with $q=1$, two queries of the function, $\ell$, are performed in each step.
To estimate the gradient sign, we only need to compare the losses and check whether $\ell(\vb{x} + \vb{u_i}) > \ell(\vb{x} - \vb{u_i})$ or $\ell(\vb{x} + \vb{u_i}) < \ell(\vb{x} - \vb{u_i})$ and multiply it with the entry-wise sign of the noise, $\vb{u_i}$.

The second adaptation involves adjusting the learning rate of ZO-Sign-SGD.
To maintain parameter values within a quantized scale, we define the adjusted learning rate as:
\begin{align} \label{eq:lr}
\eta_q = c(\eta), \quad c(\eta) = \max\{\lfloor\frac{\eta}{s}\rfloor, 1\} s. \end{align}
The parameters are updated according to the sign of the estimated gradient, the adjusted learning rate $\eta_q$, and the valid range for the quantized parameters.
All the parameters are updated with the same magnitude regardless of their true gradient magnitude.

The loss function involves FP calculations and can be typically expressed as $\ell(\vb{x}) = \ell(f(\vb{x}))$ and usually requires floating-point operations, where $f(\vb{x})$ is the output logit.
In our case, calculating the logit is quantized due to the quantized parameters of the model, while $\ell(\cdot)$ remains in its original FP form.
The logit input to $\ell(\cdot)$ is usually low dimensional compared to the dimension of $\vb{x}$.
Note that the loss usually includes softmax and cross-entropy and can be approximated efficiently on edge devices by using hash tables which include pre-computed values of the exponents and logs of the quantized values. 

Our approach uses only the sign of the loss differences of two queries at each iteration. Note that for just calculating the sign of the loss difference, an exact value of the loss is not necessary, which opens the door for potential future improvement to our method.
One possible direction to explore is using tools from human comparison feedback training, where humans are asked to compare which of two models performed better \cite{christiano2017deep}. Note that this task bears parallels to the calculation of which noisy parameters encountered a lower loss value. We defer this exploration to a future work. 

Additionally, generating the distribution of the noise, $\mathcal{B}$, may involve floating-point calculations. 
The Gaussian noise typically used is not discrete, so quantizing it still requires floating-point operations.
The discritization of the distribution is done before training and it can be used throughout the training since the valid range of the parameters remains unchanged. 
The distribution vector length is $2m+1$ which depends on the ratio between the ZO step, $\mu$ and the scale, $s$.
$\mu$ is generally small and $s$ gets larger as the quantization budget gets smaller, so the expected dimension of the distribution is small (in our empirical experiments the probability vector is up to length $20$).

When using the approximated gradient of RGE (\cref{RGE}) and calculating their signs,  one needs to discretize the distribution from which the noise is sampled. 
For this, we discretize the normal distribution $\mathcal{N}(0, \mu^2)$ by normalizing the Probability Distribution Function (PDF), $f(x) = \frac{1}{\sqrt{2\pi\mu^2}} e^{\frac{-x^2}{2\mu^2}}$.
We use the values of $f(\cdot)$ at the quantization levels to form a finite probability distribution:
$$\mathcal{D}(\mathcal{B}) = \left[\frac{f(-ms)}{\sum_{z\in\mathcal{B}}f(z)},,..., \frac{f(0)}{\sum_{z\in\mathcal{B}}f(z)},..., \frac{f(ms)}{\sum_{z\in\mathcal{B}}f(z)}\right].$$
Note that as this is a low dimensional vector, it can be hashed in a non-quantized way and efficiently used during training. 
A simpler option is to sample from a uniform distribution but the deviation from Gaussian distribution may be significant and harm performance.

\changes{Memory efficiency is a key advantage of our method.
The required memory for ZOQO update step is:
$\mathcal{M}(\vb{x}_i^q, \vb{u}_i,  \vb{x}^{\texttt{noisy}}_i, \ell(\vb{x}^+), \ell(\vb{x}^-)) = 3bd+ 2b_{\texttt{FP}},$
where $b_{\texttt{FP}}$ is the bit-width required to store FP variables.
In particular, we eliminate the need to store parameters of dimension $d$ in FP, unlike QAT, where the gradients are stored in FP, adding memory of $b_{\texttt{FP}}d$.}

\section{Experiments}
To assess the performance of our method, we evaluate it in two scenarios: (i) black-box adversarial attacks, where only input-output access to the model is available, and (ii) zero-order fine-tuning of LLMs.
We compare ZOQO with sign-based optimization methods to asses the effect of quantization.
It is important to note that our method does not have straightforward comparison benchmarks, as existing baselines rely on full-precision operations during training for gradient calculations.
Thus, we compare ZOQO against approaches that quantize the parameters either after training or during training.

\textbf{Setup for black-box adversarial attacks.}
In the case of black-box adversarial attacks, we combine sign based zero order attacks with multiple environments: (i) \textit{FP} where we use models and attacks trained with FP, (ii) \textit{Quant. Model} where the pre-trained models are quantized but attacks is performed in FP and (iii) \textit{Post Quant.} where we learn the attack in a FP environment and quantize it after learning.
We compare these approaches with our method, which does not involve full-precision computations (other than the sign of the losses).
Note that in this way, we further expose the vulnerability of deep learning quantized models to adversarial attacks where even in low resource environment the model can easily be attacked.
Additionally, robust models that are quantized after training are also exposed to threats even from an adversary with low computational resources.
 
We utilize the code provided for SignHunter \cite{al2020sign} to implement zero-order adversarial attacks.
We evaluate our method on models from the adversarial challenges for MNIST and CIFAR-10 \cite{madry2017towards}, which are both naturally and adversarially trained.
For ImageNet, we use the naturally trained TensorFlow’s Inception v3 model.

We integrate ZOQO with two methods for approximating gradient signs: SignHunter \cite{al2020sign}, which efficiently approximates the signs with noise in $\{\pm\mu\}$, and RandomSign, which employs random inefficient sign estimation.
We apply the same hyperparameters as those detailed in \cite{al2020sign}. 
The maximum number of allowed queries is 10,000.
We evaluate our method using 10,000, 1,000 and 1,000 examples from the MNIST, CIFAR-10 and 1,000 ImageNet test sets, respectively. 
We use $\ell_{\infty}$-bounded attacks with $\epsilon=0.3,12,0.05$ for MNIST, CIFAR-10, and ImageNet, respectively.
SignHunter and RandomSign both inject discrete uniform magnitude of noise into the images.
For quantization, we simply scaled the noise $\max\{s, \lfloor \frac{\mu}{s}\rfloor s\}$ so that the noise will be in the quantization levels.
As a baseline, we also compare our method with the easily quantizable Simple attack \cite{guo2019simple}, which does not require gradients to learn the attacks.
This method sequentially iterates over the coordinates and tests if adding or subtracting $\mu$ increase the loss, constructing the attack accordingly. For quantization we scale $\mu$ to be on the level of quantization.

\noindent \textbf{Setup for LLM fine-tuning.} 
Fine-tuning of LLMs may require large amount of resources. Although quantized models can achieve good performance, training them in a fully quantized way is non-trivial.
We experiment with ZO-Sign-SGD in multiple quantization environments: (i) \textit{FP} zero-order fine-tuning; (ii) \textit{Post Q.} where we quantize the trained model after the fine-tuning process; (iii) \textit{Q. post Updates} where the model is quantized after the use of non-quantized training step; and (iv) \textit{Q. Noise and Q. post update} where Gaussian noise is added to the parameters and then the parameters are quantized, the learning rate is not quantized and after each update the parameters are re-quantized.
While we do not expect our method to perform as well as these methods, which rely on full-precision operations, our results demonstrate that ZOQO manages to achieve non-trivial performance without relying on full-precision calculations. 

We evaluated our method using the ZO benchmark \cite{zhang2024revisiting} on the SST2 sentiment analysis task, fine-tuning the OPT-1.3b pre-trained model with LoRA. 
We use the same hyper-parameters as in \cite{zhang2024revisiting}. We use 20,000 update steps and 1,000 examples. 
The high dimensionality of the model makes gradient sign approximation challenging.
We apply layer-wise parameter quantization, where each layer has its own $R_{\min}$, $R_{\max}$, and scaling factor $s$.
Since the original ZO-Sign approximation in this setting is performed with Gaussian noise, we use its discrete approximation.
To handle LoRA's random initialization, we allow a wider range for parameter magnitude by extending the range to twice the initial minimal and maximal values.
Our results may differ from \cite{zhang2024revisiting} as we use the discrete ZO-Sign-SGD from \cite{liu2019signsgd}, rather than the continuous update steps involving Gaussian noise.

\begin{table}[t]
\centering
\caption{Comparison of Simple and SignHunter for adversarial attacks with 8-bit / 4-bit quantization, respectively.}\label{tab:small}
\begin{tabular}{ll cc}
\toprule
\textbf{Dataset} & \textbf{Method} & \textbf{Post quant.} & \makecell{\textbf{Quant.} \\ \textbf{attack}} $\downarrow$\\ \midrule

\multirow{4}{*}{MNIST}  
    & Simple       & 0.80 / 0.81 & 0.80 / 0.81 \\
    & SignHunter   & 0.00 / 0.00 & 0.00 / 0.00 \\
    \cmidrule{2-4}
    & Simple (adv)    & 0.93 / 0.05 & 0.92 / 0.03 \\
    & SignHunter (adv)& 0.89 / 0.07 & 0.89 / 0.00 \\ \midrule

\multirow{4}{*}{CIFAR-10} 
    & Simple       & 0.51 / 0.95 & 0.51 / 0.24 \\
    & SignHunter   & 0.08 / 0.03 & 0.09 / 0.02 \\
    \cmidrule{2-4}
    & Simple (adv)    & 0.94 / 0.99 & 0.96 / 0.82 \\
    & SignHunter (adv)& 0.53 / 0.33 & 0.56 / 0.57 \\

\bottomrule
\end{tabular}
\vspace{-0.2cm}
\end{table}

\noindent\textbf{Memory Efficiency.}
We also simulate memory usage for a single update step using a toy model. The memory calculation includes all tensors allocated during training, determined by their number of elements and their precision.
The experiment employs a fully connected network with three layers, an input dimension of $784$, an output dimension of $10$, and a batch size of $16$.
We compare three scenarios: (1) FP training, (2) QAT, where forward passes use quantized weights, and backward passes use FP, and (3) ZOQO, where updates are entirely quantized. For the quantized parameters we use $b=8$ bits, and the calculations include memory allocated for FP data.
As shown \cref{tab:memory}, ZOQO achieves a significant reduction in memory usage compared to FP training and QAT.
\begin{table}[t]
    \centering
\caption{Accuracy of ZO-Sign-SGD \cite{liu2019signsgd}, \changes{ZO-Adam \cite{chen2019zo} and ZOQO} fine-tuning of OPT-1.3b model with LoRA on SST2. \textit{Q. post updates} refers to quantizing the parameters after each FP update and \textit{Q. noise} refers to quantization of the noisy parameters, with non-quantized learning rate.
}
    \label{tab:ft_llm}
    \begin{tabular}{ccccc}
        \toprule
        \textbf{Method} & \textbf{Bits} & \textbf{Post-Q.} & \makecell{\textbf{Q. post}\\ \textbf{updates}} & \makecell{\textbf{Q. noise \&} \\\textbf{Q. post updates}} $\uparrow$ \\
        \midrule
        \textbf{ZO-Sign} & $b=8$ & 91.06 & 88.41 & 79.82 \\
        (91.28) & $b=4$ & 51.49 & 56.77 & 53.32 \\
        \midrule
        \textbf{ZO-Adam} & $b=8$ & 91.63 & 85.10 & 61.81 \\
        (92.32) & $b=4$ & 51.49 & 53.55 & 54.70 \\
        \midrule
        \textbf{ZOQO}            & $\textbf{b=(8/4)}$ & \multicolumn{3}{c}{\textbf{89.68 \hspace{0.3cm}/ \hspace{0.3cm}64.34}} \\
        \bottomrule
    \end{tabular}
    \vspace{-0.5cm}
\end{table}

\begin{table}[t]
    \centering
    \caption{\changes{A simulation results of peak memory usage of a single update of a toy model. Note the significant benefit of using ZOQO.}}
    \label{tab:memory}
            \begin{tabular}{lccc}
    \toprule
         & \textbf{FP} & \textbf{QAT} & \textbf{ZOQO} $\downarrow$  \\ 
         \midrule
         Memory Peak [MB] & 903.71 & 583.83 & \textbf{371.21}  \\ 
        \bottomrule
    \end{tabular}
    \vspace{-0.6cm}
\end{table}

\noindent\textbf{Results. }
\cref{tab:adv_cifar_mnist} presents the results for black-box adversarial attacks.
Generally, models that are more aggressively quantized tend to exhibit lower failure rates, as attacking models with reduced precision is inherently easier due to the coarser parameter space.
Despite the increased difficulty of maintaining performance in low-precision environments, ZOQO is remarkably resilience, often causing only minimal harm to the overall performance.
In several cases, the results for ZOQO are comparable to full-precision attacks on quantized models and attacks performed on models that are quantized after learning.
Notably, aggressive quantization of the MNIST robustly trained model at $b=4$ bits significantly increases its vulnerability to adversarial attacks, which diminishes its intended robustness.
\cref{tab:small} compares to Simple attack \cite{guo2019simple}.
The quantized version of Simple attack underperforms compared to ZOQO with SignHunter, emphasizing the effectiveness of quantized sign-based optimization.
For ImageNet, \cref{tab:adv_imagenet} exhibits that ZOQO suffers only minor performance degradation compared to FP models and attacks.
The robustness of ZOQO across different datasets highlights its general applicability to a variety of adversarial settings, making it a viable method for constrained optimization environments.

Finally, fine-tuning LLMs using LoRA, as presented in \cref{tab:ft_llm}, further demonstrates the practical efficacy of ZOQO.
At $b=8$ bits, while ZOQO does incur some performance drop compared to post-training quantization, it outperforms baselines that quantize the model after each update step and those that apply quantization to the noise but leave the learning rate unquantized.
For $b=4$ bits, the impact of post-training quantization is more severe, resulting in a substantial loss of performance. 
However, ZOQO achieves the highest accuracy across all baselines, showcasing its potential to handle highly quantized environments more effectively.

\noindent\textbf{Ablation and Analysis.} The baselines presented in \cref{tab:ft_llm} can be considered as an ablation of our method.
Specifically, the baselines involving quantization of parameters after each non-quantized update and quantization of noisy parameters without applying quantization to the learning rate represent partial and inefficient implementations of our approach. 
The results clearly demonstrate that these additional inefficient steps, such as non-quantized updates and non-quantized noise, lead to a noticeable degradation in performance.

Overall, our results show that ZOQO, which eliminates the need for full-precision operations during training, achieves comparable performance to methods that rely on full-precision computations, with minor performance degradation in some cases.
ZOQO proves effective for adversarial attacks and fine-tuning in resource-constrained environments where both operations and parameters are quantized.

\section{Conclusions and Future Work}
In this paper, we introduced a quantized zero-order optimization method that operates with limited precision, using only quantized parameters and gradient approximations.
We applied our method to black-box adversarial attacks and zero-order fine-tuning of LLMs, demonstrating that it is possible to achieve minimal harm in performance without relying on full-precision operations.
Our performance results show that ZOQO can successfully handle challenging high-dimensional optimization tasks, making it a promising approach for scenarios where FP operations are restricted or costly.

ZOQO opens the door to many follow-ups.
Future work includes developing a rigorous convergence theory for ZOQO in convex and bounded cases, focusing on establishing clear theoretical guarantees and addressing potential challenges. This analysis will help determining the precise conditions under which ZOQO converges efficiently.
We will also explore ZOQO's applicability in other machine learning areas, including reinforcement learning and distributed optimization. 
Further work will employ advanced quantization techniques, such as adaptive quantization, which could further boost performance.
Additionally, examining methods to estimate the sign of the loss differences without the need for full loss calculations will help minimizing the computational overhead. A possible direction might be taking inspiration from human feedback models \cite{christiano2017deep}.

\clearpage
\bibliographystyle{IEEEbib}
\bibliography{quant_learn}

\end{document}